\documentclass{article}

\usepackage[preprint]{neurips_2019}

\usepackage[utf8]{inputenc} 
\usepackage[T1]{fontenc}    
\usepackage{hyperref}       
\usepackage{url}            

\usepackage{booktabs}       
\usepackage{amsmath}
\usepackage{amssymb}
\usepackage{amsfonts}       
\usepackage{caption}
\usepackage{nicefrac}       
\usepackage{microtype}      
\usepackage{graphicx}
\usepackage{subcaption}
\usepackage{lipsum}
\usepackage{pifont}
\usepackage{xfrac}
\usepackage{tikz}
\usepackage{xcolor}
\usepackage{enumitem}
\usepackage{wrapfig}

\newcommand{\gray}[1]{{\color{gray}#1}}
\newcommand{\black}[1]{{\color{black}#1}}

\usepackage{ulem} 
\usepackage{caption} 
\setlist{nolistsep}

\usetikzlibrary{arrows,shapes.geometric,positioning}

\hypersetup{colorlinks=true,citecolor=blue}

\newcommand{\cmark}{\ding{51}}%
\newcommand{\xmark}{\ding{55}}%
\newcommand{\eos}{\langle \mathrm{EOS} \rangle}

\DeclareMathOperator*{\argmax}{argmax}

\title{KERMIT: \\ Generative Insertion-Based Modeling for Sequences}

\author{%
  William Chan\thanks{Equal contribution. WC initiated the KERMIT project for machine translation, implemented the corresponding code and experiments (Section \ref{sec:machinetranslation}) and advised the project. NK proposed using the same model for text generation, implemented and evaluated different monolingual pre-training approaches, and conducted all representation learning experiments (Section \ref{sec:representationlearning}). KG proposed using KERMIT as a zero-shot QA model and conducted all associated experiments (Section \ref{sec:clozeqa}); he also co-developed KERMIT's training and inference infrastructure. MS developed the mathematical formalism for the model (Section \ref{sec:kermit}) and assisted in the implementation of KERMIT for translation. JU helped conceive the initial idea and advised the project.}\hspace{1.75mm}$^1$,
  Nikita Kitaev\footnotemark[1]\hspace{1.75mm}$^{1,3}$,
  Kelvin Guu\footnotemark[1]\hspace{1.75mm}$^2$,
  Mitchell Stern\footnotemark[1]\hspace{1.75mm}$^{1,3}$,
  Jakob Uszkoreit$^1$ \\
  $^1$Google Research, Brain Team \\
  $^2$Google Research, AI Language Team \\
  $^3$University of California, Berkeley \\
  \texttt{\{williamchan,kguu,usz\}@google.com} \\
  \texttt{\{kitaev,mitchell\}@berkeley.edu} \\
}

\begin{document}

\maketitle

\begin{abstract}
We present KERMIT, a simple insertion-based approach to generative modeling for sequences and sequence pairs. KERMIT models the joint distribution and its decompositions (i.e., marginals and conditionals) using a single neural network and, unlike much prior work, does not rely on a prespecified factorization of the data distribution. During training, one can feed KERMIT paired data $(x, y)$ to learn the joint distribution $p(x, y)$, and optionally mix in unpaired data $x$ or $y$ to refine the marginals $p(x)$ or $p(y)$. During inference, we have access to the conditionals $p(x \mid y)$ and $p(y \mid x)$ in both directions. We can also sample from the joint distribution or the marginals. The model supports both serial fully autoregressive decoding and parallel partially autoregressive decoding, with the latter exhibiting an empirically logarithmic runtime. We demonstrate through experiments in machine translation, representation learning, and zero-shot cloze question answering that our unified approach is capable of matching or exceeding the performance of dedicated state-of-the-art systems across a wide range of tasks without the need for problem-specific architectural adaptation.
\end{abstract}

\section{Introduction}
\label{sec:introduction}

Neural sequence models \citep{sutskever-nips-2014,cho-emnlp-2014} have been successfully applied to many conditional generation applications, including machine translation \citep{bahdanau-iclr-2015,luong-emnlp-2015}, speech recognition \citep{chan-icassp-2016,bahdanau-icassp-2016}, speech synthesis \citep{oord-arxiv-2016,wang-interspeech-2017} and image captioning \citep{vinyals-cvpr-2015,xu-icml-2015}.
Much of the prior work in this area follows the seq2seq encoder-decoder paradigm, where an encoder builds a representation of an observed sequence $x$, and a decoder gives the conditional output distribution $p(y \mid x)$ according to a predetermined factorization, usually left-to-right.

While effective for straightforward conditional generation, such an approach is inflexible and cannot readily be applied to other inference tasks such as non-left-to-right generation or infilling. In this work, we present a more general approach called Kontextuell Encoder Representations Made by Insertion Transformations, or KERMIT for short. KERMIT is a simple architecture that directly models the joint distribution $p(x, y)$ and its decompositions (such as the marginals $p(x)$ and $p(y)$ and the conditionals $p(y \mid x)$ and $p(x \mid y)$) in a unified manner. In contrast with traditional seq2seq models, KERMIT does not rely on a prespecified factorization, but is instead able to condition on whatever information is available and infer what remains.

During training, we present KERMIT with paired data $(x, y)$ to learn the joint, and can optionally mix in unpaired data $x$ or $y$ to refine the marginals in a semi-supervised setting. At test time, a single KERMIT model can be used for conditional inference in either direction by restricting the output distribution to $p(x \mid y)$ or $p(y \mid x)$ as required. We can also generate paired samples from the joint distribution $(x, y) \sim p(x, y)$, or unpaired samples from the marginals $x \sim p(x)$ or $y \sim p(y)$.

\begin{figure}[t]
\small
\centering
\begin{tikzpicture}[xscale=1.7,yscale=0.8]
\tikzset{every node/.style={anchor=base}, every path/.style={->,>=triangle 45,shorten >=4pt}}
\foreach \word [count=\i] in {courses,proved,popular,[SEP],Die,sehr,beliebt,[SEP]} \node at (\i, 0) {\word};
\draw (0.5, 0) -- ++(0, 1) node {The};
\draw (2.5, 0) -- ++(0, 1) node {quite};
\draw (5.5, 0) -- ++(0, 1) node {\{Kurse, waren\}};
\end{tikzpicture}
\caption{An example of the KERMIT insertion objective for the English $\leftrightarrow$ German translation pair ``The courses proved quite popular'' $\leftrightarrow$ ``Die Kurse waren sehr beliebt''. The model is trained to predict the set of words that need to be inserted at each location. By incurring a loss on both sides, our system learns a fully generative model of the joint distribution over $(x, y)$ pairs, and can accommodate arbitrary generation orders.}
\label{fig:insertion-objective}
\end{figure}
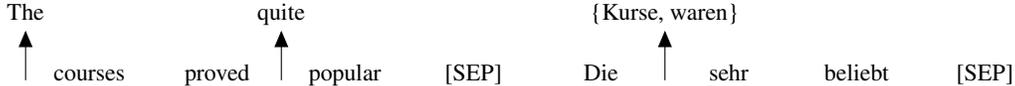

KERMIT uses a simple architecture and is easy to implement. It does not have a separate encoder and decoder, nor does it require causality masks. In our implementation, KERMIT consists of a single Transformer decoder stack \citep{vaswani-nips-2017}. The model is trained to insert the missing tokens into any partially-complete sequence, as shown in Figure~\ref{fig:insertion-objective}. We describe the implementation in more detail in Section~\ref{sec:kermit}.

We apply KERMIT to a diverse set of tasks, finding that our unified approach is capable of matching or exceeding the performance of dedicated state-of-the-art systems without the need for problem-specific components. We first apply KERMIT to machine translation, where the inputs and outputs are parallel sentence pairs.
Then, like its friends ELMo \citep{peters-arxiv-2018}, BERT \citep{devlin-naacl-2019}, and ERNIE \citep{sun-arxiv-2019}, we can also use KERMIT for self-supervised representation learning for use in downstream NLP tasks.
Finally, we apply KERMIT to a zero-shot cloze question-answering task demonstrating the infilling capabilities of the model.
Table \ref{tab:comparison} summarizes our results on all three tasks compared to other highly tuned models: Transformer, BERT, GPT and GPT-2.


\begin{table}[t]
\small
\centering
\resizebox{\textwidth}{!}{
\begin{tabular}{lccc}
\toprule
& \bfseries Machine Translation & \bfseries  Representation Learning & \bfseries Cloze Question Answering \\
& En $\rightarrow$ De (BLEU) & GLUE & Zero-shot SQuAD (F1) \\
\midrule
Autoregressive (Transformer, GPT, GPT-2) & 27.3$^a$ & 72.8$^b$ & 16.6 \\
Masking (BERT) & N/A\phantom{$^c$} & 80.5$^c$ & 18.9 \\
Insertion (KERMIT -- Our Work) & 27.8\phantom{$^c$} & 79.8\phantom{$^c$} & 30.3 \\
\bottomrule
\end{tabular}%
} 
\caption{The KERMIT architecture works well for three categories of tasks: machine translation, representation learning, and zero-shot cloze question answering. $^a$\citet{vaswani-nips-2017} $^b$\citet{radford-openai-2018} $^c$\citet{devlin-naacl-2019}}
\label{tab:comparison}
\end{table}
\section{Background}
\label{sec:background}

In this section, we define some notation and give a brief review of existing sequence models, including autoregressive left-to-right models \citep{sutskever-nips-2014,cho-emnlp-2014} and masked language models \citep{devlin-naacl-2019}.

\subsection{Autoregressive Left-to-Right Models}
Let $\mathcal{X}$ and $\mathcal{Y}$ be the set of all input and output sequences, respectively. In a standard sequence-to-sequence task, we are presented with training data consisting of sequence pairs $(x, y) \in \mathcal{X} \times \mathcal{Y}$, e.g.\ parallel translations, and we aim to learn the conditional distribution $p(y \mid x)$.
Traditional autoregressive models \citep{sutskever-nips-2014,cho-emnlp-2014} use a left-to-right factorization, decomposing the distribution as a chain of predictions conditioning on the input $x$ and prefixes $y_{<t}$:
\begin{align}
    p(y \mid x) = \prod_t p(y_t \mid x, y_{<t}) .
\end{align}
This structure is also used for unconditional sequence tasks such as language modeling where the goal is to learn an unconditional output distribution on its own. A left-to-right factorization is convenient because it allows for exact log-likelihood computation, thereby permitting efficient maximum likelihood estimation.
It also leads to simple approximate inference algorithms such as greedy decoding
\begin{align}
    \hat{y}_t = \argmax_y p(y \mid x, \hat{y}_{<t})
\end{align}
or beam search over sets of multiple hypotheses.

However, there are some drawbacks to the autoregressive approach. First, in the case of conditional generation, it cannot handle situations where the input $x$ is only partially observed. Second, since it utilizes a fixed left-to-right factorization, it cannot be used for other inference tasks like infilling where generation is not monotonic. Moreover, standard inference algorithms require $n$ generation steps to generate $n$ tokens, which could be a bottleneck in end-use applications.

\subsection{Masked Language Models}
Masked Language Models (MLMs) \citep{devlin-naacl-2019} comprise another class of models targeting the unconditional setting. For MLMs, a partial canvas $x_s \subseteq x$ is observed where some of the tokens in $x$ have been masked out, and the objective is to recover $x$ from $x_s$.
For example, for a ground truth canvas $x^* = (A, B, C, D, E)$ and a partial canvas $x^*_s = (A, \_,  C, D, \_)$, the model should learn to replace the second blank with $B$ and the last blank with $E$.
The model outputs an independent prediction at each position, and its objective is to maximize $p(x \mid x_s)$.


Because the exact locations of the slots are known in $x_s$, the model does not need to predict where the missing items are located, but only what they should be.
Consequently, the model is not immediately suitable for generation, as the canvas size needs to be fixed during inference and cannot change over time (i.e., $|x_s| = |x|$). MLMs have been successfully applied in self-supervised representation learning settings, leading to strong results on downstream language tasks \citep{devlin-naacl-2019}.
\section{KERMIT}
\label{sec:kermit}

In this section we propose KERMIT, a novel insertion-based generative model. Unlike the prior work mentioned in Section~\ref{sec:background}, KERMIT does not have the rigid construction of modeling the target sequence given some fully observed source sequence, nor does it assume a left-to-right factorization (and generation order) of the output sequence. To motivate and arrive at our model, we formalize then extend a recent insertion-based conditional modeling framework proposed by \citet{stern-icml-2019}.

We begin with the unconditional setting. In order to model sequences without requiring a fixed factorization or imposing constraints on the order of generation, we make use of a framework in which sequences are constructed via insertion operations. Given a sequence $x = (x_1, \dots, x_n)$ and a generation order $z$ represented as a permutation of the indices $\{1, \dots, n\}$, we define the corresponding sequence $((c^z_1, l^z_1), \dots, (c^z_n, l^z_n))$ of insertion operations which produces $x$ according to order $z$. Here, $c^z_i \in \mathcal{C}$ is an element of the vocabulary and $1 \le l^z_i \le i$ is an insertion location relative to the current hypothesis. For example, if constructing the sequence $(A, B, C)$ as $() \to (C) \to (A, C) \to (A, B, C)$, we would have $z = (3, 1, 2)$ with $(c^z_1, l^z_1) = (C, 1), (c^z_2, l^z_2) = (A, 1), (c^z_3, l^z_3) = (B, 2)$.

Next let $(x^{z,i}_1, \dots, x^{z,i}_i)$ denote the subsequence of $x$ corresponding to the (ordered) extraction of the elements at indices $\{z_1, \dots, z_i\}$. This is the partial output at iteration $i$. Note that this will be the same for all permutations $z$ with the same unordered set of indices in the first $i$ positions. For the example above for instance, we have $(x^{z,2}_1, x^{z,2}_2) = (A, C)$.

Armed with these definitions, we can now write out $p(x)$ as a marginalization over all possible orders $z \in S_n$ for sequence length $n$, where $S_n$ denotes the set of all permutations on $n$ elements:
\begingroup
\allowdisplaybreaks
\begin{align}
p(x)
& = \sum_{z \in S_n} p(x, z) \\
& = \sum_{z \in S_n} p(z) p(x \mid z) \\
& = \sum_{z \in S_n} p(z) \prod_{i=1}^n p((c^z_i, l^z_i) \mid (c^z_1, l^z_1), \dots, (c^z_{i-1}, l^z_{i-1})) \\
& = \sum_{z \in S_n} p(z) \prod_{i=1}^n p((c^z_i, l^z_i) \mid x^{z,i-1}_{1:i-1}) ,
\end{align}
\endgroup
where the last line encodes the Markov assumption that the order of insertions leading to a given canvas is not important, just the result. Typically we will use a uniform prior over permutations for $p(z)$, though other options are available, such as the balanced binary tree prior described by \citet{stern-icml-2019}.

\subsection{Learning}

Although exact computation of the log-likelihood is intractable due to the marginalization over the generation order $z$, we can lower bound the log-likelihood using Jensen's inequality via
\begin{align}
\log p(x)
& = \log \sum_{z \in S_n} p(z) p(x \mid z) \\
& \ge \sum_{z \in S_n} p(z) \log p(x \mid z) \quad =: \mathcal{L}(x). \\
\intertext{Substituting in our expression for $p(x \mid z)$ from above, we have}
\mathcal{L}(x)
& = \sum_{z \in S_n} p(z) \log \prod_{i=1}^n p((c^z_i, l^z_i) \mid x^{z,i-1}_{1:i-1}) \\
& = \sum_{z \in S_n} p(z) \sum_{i=1}^n \log p((c^z_i, l^z_i) \mid x^{z,i-1}_{1:i-1}) .
\intertext{Next we interchange the summations and break the permutation $z$ down into $(z_1, \dots, z_{i-1})$ corresponding to previous insertions, $z_i$ corresponding to the next insertion, and $(z_{i+1}, \dots, z_n)$ corresponding to future insertions, giving}
\mathcal{L}(x)
& = \sum_{i=1}^n \sum_{z \in S_n} p(z) \log p((c^z_i, l^z_i) \mid x^{z,i-1}_{1:i-1}) \\
& = \sum_{i=1}^n \sum_{z_{1:i-1}} \sum_{z_i} \sum_{z_{i+1:n}} p(z) \log p((c^z_i, l^z_i) \mid x^{z,i-1}_{1:i-1}) \\
& = \sum_{i=1}^n \sum_{z_{1:i-1}} p(z_{1:i-1}) \sum_{z_i} p(z_i \mid z_{1:i-1}) \log p((c^z_i, l^z_i) \mid x^{z,i-1}_{1:i-1}) \sum_{z_{i+1:n}} p(z_{i+1:n} \mid z_{1:i}) \\
& = \sum_{i=1}^n \sum_{z_{1:i-1}} p(z_{1:i-1}) \sum_{z_i} p(z_i \mid z_{1:i-1}) \log p((c^z_i, l^z_i) \mid x^{z,i-1}_{1:i-1}) ,
\end{align}


where the simplification in the last line follows from the fact that $\sum_{z_{i+1:n}} p(z_{i+1:n} \mid z_{1:i}) = 1$.

From here, we can multiply and divide the outer sum by $n$ to turn it into a mean, then arrive at the following simple sampling procedure to compute an unbiased estimate of our lower bound $\mathcal{L}(x)$ on the log-likelihood for a single example:
\begin{enumerate}
\item Sample a generation step $i \sim \mathrm{Uniform}([1, n])$.
\item Sample a partial permutation $z_{1:i-1} \sim p(z_{1:i-1})$ for the first $i-1$ insertions.
\item Compute a weighted sum over the next-step losses $\log p((c^z_i, l^z_i) \mid x^{z,i-1}_{1:i-1})$ scaled by the weighting distribution $p(z_i \mid z_{1:i-1})$ and the sequence length $n$.
\end{enumerate}


\subsection{Inference}
Using this model, inference can be autoregressive via greedy decoding
\begin{align}
    (\hat c, \hat l) = \argmax_{c, l} p(c, l | \hat{x}_t)
\end{align}
or partially autoregressive via parallel decoding
\begin{align}
    \hat{c}_{l} = \argmax_{c} p(c \mid l, \hat{x}_t) .
\end{align}
In the case of parallel decoding, we perform simultaneous insertions at all non-finished slots. If we use a balanced binary tree prior for $p(z)$ \citep{stern-icml-2019}, we can even achieve an empirical runtime of $\approx \log_2 n$ iterations to generate $n$ tokens. One key advantage of insertion-based models over MLMs is that the output canvas can dynamically grow in size, meaning the length does not need to be chosen before the start of generation.

\subsection{Pairs of Sequences}
Thus far, we have discussed KERMIT for single sequences. We can easily extend KERMIT to pairs of sequences by directly modeling $(x, y)$ as a concatenation of two sequences, $(x, y) = (x_1, \dots, x_n, y_1, \dots, y_m)$. For example, let our first sequence be $x = (A, B, C, \eos)$ and our second sequence be $y = (A', B', C', D', E', \eos)$. The concatenated sequence would then be $(x, y) = (A, B, C, \eos, A', B', C', D', E', \eos)$.
With this approach, we can model pairs of sequences as if they were single sequences. Moreover, unlike seq2seq, our model is symmetric with regards to its treatment of the source and target, making it a strong candidate for extensions to multimodal data settings in future work.

By keeping our architecture order-agnostic and marginalizing over all possible orders in our training objective, KERMIT is able to learn the joint distribution and all its decompositions, including the marginals $p(x)$ and $p(y)$ and conditionals $p(y \mid x)$ and $p(x \mid y)$. We can also perform targeted training. More explicitly, if the model is provided with a canvas that fully contains $x$ or $y$, then it will learn a conditional distribution. If the model is provided with an example where $x$ or $y$ is empty, then it will learn the opposing marginal distribution.

\subsection{Model}

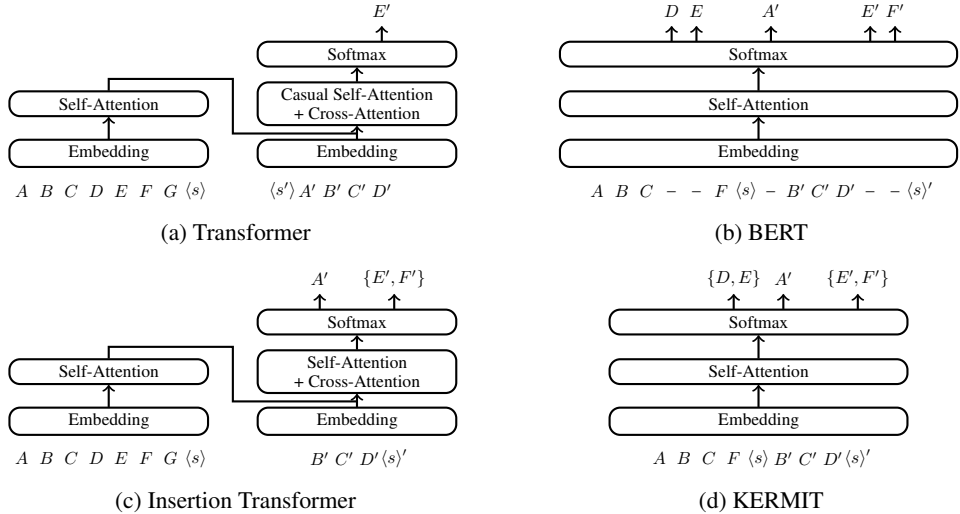
\begin{figure}
\centering
\begin{subfigure}{0.5\textwidth}
\centering
\begin{tikzpicture}[node distance=1cm, scale=0.66, every node/.style={transform shape}]
\tikzstyle{layer} = [rectangle, thick, rounded corners, minimum width=4cm, minimum height=0.5cm, align=center, draw=black]
\tikzstyle{symbol} = [align=center, minimum height=0.6cm]
\tikzstyle{arrow} = [thick,->]

\node (d2) [layer] at (0, 0) {Softmax};
\node (d1) [layer, below of=d2] {Casual Self-Attention \\ + Cross-Attention};
\node (d0) [layer, below of=d1] {Embedding};

\node (e0) [layer] at (-5,-2.0) {Embedding};
\node (e1) [layer, above of=e0] {Self-Attention};

\node[symbol] at (-1.5, -2.75) {$\langle s' \rangle$};
\node[symbol] at (-1.0, -2.75) {$A'$};
\node[symbol] at (-0.5, -2.75) {$B'$};
\node[symbol] at (0.0, -2.75) {$C'$};
\node[symbol] at (0.5, -2.75) {$D'$};

\node[symbol] at (-6.75, -2.75) {$A$};
\node[symbol] at (-6.25, -2.75) {$B$};
\node[symbol] at (-5.75, -2.75) {$C$};
\node[symbol] at (-5.25, -2.75) {$D$};
\node[symbol] at (-4.75, -2.75) {$E$};
\node[symbol] at (-4.25, -2.75) {$F$};
\node[symbol] at (-3.75, -2.75) {$G$};
\node[symbol] at (-3.25, -2.75) {$\langle s \rangle$};

\node (ye) [symbol] at (0.5, 0.875) {$E'$};

\draw[arrow] (d0) -- (d1);
\draw[arrow] (d1) -- (d2);

\draw[arrow] (e0) -- (e1);
\draw[thick] (e1.north) -- (-5,-0.5) -- (-2.5,-0.5) -- (-2.5,-1.6) -- (0.0,-1.6);

\draw[arrow] (d2.north) ++(0.5,0.0) -- (ye);
\end{tikzpicture}
\caption{Transformer}
\label{fig:transformer}
\end{subfigure}%
\begin{subfigure}{0.5\textwidth}
\centering
\begin{tikzpicture}[node distance=1cm, scale=0.66, every node/.style={transform shape}]
\tikzstyle{layer} = [rectangle, thick, rounded corners, minimum width=8cm, minimum height=0.5cm,text centered, draw=black]
\tikzstyle{symbol} = [align=center, minimum height=0.6cm]
\tikzstyle{arrow} = [thick,->]

\node (d2) [layer] at (0, 0) {Softmax};
\node (d1) [layer, below of=d2] {Self-Attention};
\node (d0) [layer, below of=d1] {Embedding};

\node[symbol] at (-3.25, -2.75) {$A$};
\node[symbol] at (-2.75, -2.75) {$B$};
\node[symbol] at (-2.25, -2.75) {$C$};
\node[symbol] at (-1.75, -2.75) {$\_$};
\node[symbol] at (-1.25, -2.75) {$\_$};
\node[symbol] at (-0.75, -2.75) {$F$};
\node[symbol] at (-0.25, -2.75) {$\langle s \rangle$};
\node[symbol] at (0.25, -2.75) {$\_$};
\node[symbol] at (0.75, -2.75) {$B'$};
\node[symbol] at (1.25, -2.75) {$C'$};
\node[symbol] at (1.75, -2.75) {$D'$};
\node[symbol] at (2.25, -2.75) {$\_$};
\node[symbol] at (2.75, -2.75) {$\_$};
\node[symbol] at (3.25, -2.75) {$\langle s \rangle'$};

\node (xd) [symbol] at (-1.75, 0.875) {$D$};
\node (xe) [symbol] at (-1.25, 0.875) {$E$};

\node (ya) [symbol] at (0.25, 0.875) {$A'$};
\node (ye) [symbol] at (2.25, 0.875) {$E'$};
\node (yf) [symbol] at (2.75, 0.875) {$F'$};

\draw[arrow] (d0) -- (d1);
\draw[arrow] (d1) -- (d2);

\draw[arrow] (d2.north) ++(-1.75,0.0) -- (xd);
\draw[arrow] (d2.north) ++(-1.25,0.0) -- (xe);

\draw[arrow] (d2.north) ++(0.25,0.0) -- (ya);
\draw[arrow] (d2.north) ++(2.25,0.0) -- (ye);
\draw[arrow] (d2.north) ++(2.75,0.0) -- (yf);
\end{tikzpicture}
\caption{BERT}
\label{fig:bert}
\end{subfigure}
\begin{subfigure}{0.5\textwidth}
\vspace{.5em}
\centering
\begin{tikzpicture}[node distance=1cm, scale=0.66, every node/.style={transform shape}]
\tikzstyle{layer} = [rectangle, thick, rounded corners, minimum width=4cm, minimum height=0.5cm, align=center, draw=black]
\tikzstyle{symbol} = [align=center, minimum height=0.6cm]
\tikzstyle{arrow} = [thick,->]

\node (d2) [layer] at (0, 0) {Softmax};
\node (d1) [layer, below of=d2] {Self-Attention \\ + Cross-Attention};
\node (d0) [layer, below of=d1] {Embedding};

\node (e0) [layer] at (-5,-2.0) {Embedding};
\node (e1) [layer, above of=e0] {Self-Attention};

\node[symbol] at (-6.75, -2.75) {$A$};
\node[symbol] at (-6.25, -2.75) {$B$};
\node[symbol] at (-5.75, -2.75) {$C$};
\node[symbol] at (-5.25, -2.75) {$D$};
\node[symbol] at (-4.75, -2.75) {$E$};
\node[symbol] at (-4.25, -2.75) {$F$};
\node[symbol] at (-3.75, -2.75) {$G$};
\node[symbol] at (-3.25, -2.75) {$\langle s \rangle$};

\node[symbol] at (-0.75, -2.75) {$B'$};
\node[symbol] at (-0.25, -2.75) {$C'$};
\node[symbol] at (0.25, -2.75) {$D'$};
\node[symbol] at (0.75, -2.75) {$\langle s \rangle'$};

\node (ya) [symbol] at (-0.75, 0.875) {$A'$};
\node (yef) [symbol] at (0.75, 0.875) {$\{E', F'\}$};

\draw[arrow] (d0) -- (d1);
\draw[arrow] (d1) -- (d2);

\draw[arrow] (e0) -- (e1);
\draw[thick] (e1.north) -- (-5,-0.5) -- (-2.5,-0.5) -- (-2.5,-1.6) -- (0.0,-1.6);

\draw[arrow] (d2.north) ++(-0.75,0.0) -- (ya);
\draw[arrow] (d2.north) ++(0.75,0.0) -- (yef);
\end{tikzpicture}
\caption{Insertion Transformer}
\label{fig:itransformer}
\end{subfigure}%
\begin{subfigure}{0.5\textwidth}
\vspace{.5em}
\centering
\begin{tikzpicture}[node distance=1cm, scale=0.66, every node/.style={transform shape}]
\tikzstyle{layer} = [rectangle, thick, rounded corners, minimum width=6cm, minimum height=0.5cm,text centered, draw=black]
\tikzstyle{symbol} = [align=center, minimum height=0.6cm]
\tikzstyle{arrow} = [thick,->]

\node (d2) [layer] at (0, 0) {Softmax};
\node (d1) [layer, below of=d2] {Self-Attention};
\node (d0) [layer, below of=d1] {Embedding};

\node[symbol] at (-2.0, -2.75) {$A$};
\node[symbol] at (-1.5, -2.75) {$B$};
\node[symbol] at (-1.0, -2.75) {$C$};
\node[symbol] at (-0.5, -2.75) {$F$};
\node[symbol] at (0.0, -2.75) {$\langle s \rangle$};
\node[symbol] at (0.5, -2.75) {$B'$};
\node[symbol] at (1.0, -2.75) {$C'$};
\node[symbol] at (1.5, -2.75) {$D'$};
\node[symbol] at (2.0, -2.75) {$\langle s \rangle'$};

\node (ya) [symbol] at (0.5, 0.875) {$A'$};
\node (yef) [symbol] at (2.0, 0.875) {$\{E', F'\}$};
\node (xde) [symbol] at (-0.5, 0.875) {$\{D, E\}$};

\draw[arrow] (d0) -- (d1);
\draw[arrow] (d1) -- (d2);

\draw[arrow] (d2.north) ++(0.5,0.0) -- (ya);
\draw[arrow] (d2.north) ++(2.0,0.0) -- (yef);
\draw[arrow] (d2.north) ++(-0.5,0.0) -- (xde);
\end{tikzpicture}
\caption{KERMIT}
\label{fig:kermit}
\end{subfigure}
\caption{Diagram of various models. The Transformer (\subref{fig:transformer}) model predicts the next right token given the left context. The BERT (\subref{fig:bert}) model predicts what is missing in the blank slots given the context. The Insertion Transformer (\subref{fig:itransformer}) model predicts where and what is missing given the context. The KERMIT (\subref{fig:kermit}) model is an generalization of (\subref{fig:itransformer}) where the context is over multiple sequences.}
\label{fig:model_comparison}
\end{figure}

We implement KERMIT as a single Transformer decoder stack \citep{vaswani-nips-2017}, without any form of causal masking. The full self-attention mechanism allows the model to capture any relationships between the input canvas and the predicted insertion operations with a constant number of operations. We follow \citet{stern-icml-2019} and model the (content, location) distribution $p(c, l)$ as a factorized distribution $p(c, l) = p(c \mid l)p(l)$, where $p(c \mid l)$ is the standard Transformer softmax over the vocabulary, and a $p(l)$ is a softmax over the locations. Figure \ref{fig:model_comparison} visualizes the differences between a standard Transformer \citep{vaswani-nips-2017}, BERT \citep{devlin-naacl-2019}, Insertion Transformer \citep{stern-icml-2019} and KERMIT.

\section{Experiments}
We perform experiments with KERMIT on the tasks of machine translation, self-supervised representation learning, and zero-shot cloze question answering.

\subsection{Machine Translation}
\label{sec:machinetranslation}

\begin{table}[t]
\centering
\small
\resizebox{\columnwidth}{!}{%
\begin{tabular}{lcccc}
\toprule
\bfseries Model & $\leftrightarrow$ & En $\rightarrow$ De & De $\rightarrow$ En & Iterations \\
\midrule
Autoregressive \\
\quad Transformer \citep{vaswani-nips-2017} & \xmark & 27.3 & & $n$ \\
\quad Transformer (Our Implementation) & \xmark & 27.8 & 31.2 & $n$ \\
\midrule
Non-Autoregressive \\
\quad NAT \citep{gu-iclr-2018} & \xmark & 17.7 & 21.5 & $1$ \\
\quad Iterative Refinement \citep{lee-emnlp-2018} & \xmark & 21.6 & 25.5 & $10$ \\
\quad Blockwise Parallel \citep{stern-neurips-2018} & \xmark & 27.4 & & $\approx n / 5$ \\
\quad Insertion Transformer \citep{stern-icml-2019} & \xmark & 27.4 & & $\approx \log_2 n \ll 10$ \\
\midrule
KERMIT \\
\quad Unidirectional ($p(y \mid x)$ or $p(x \mid y)$) & \xmark & 27.8 & 30.7 & $\approx \log_2 n \ll 10$ \\
\quad Bidirectional ($p(y \mid x)$ and $p(x \mid y)$) & \cmark & 27.2 & 27.6 & $\approx \log_2 n \ll 10$ \\
\quad Joint ($p(x, y)$) & \cmark & 25.6 & 27.4 & $\approx \log_2 n \ll 10$ \\
\quad\quad + Marginal Refining ($p(x)$ and $p(y)$) & \cmark & 25.8 & 28.6 & $\approx \log_2 n \ll 10$ \\
\quad\quad\quad $\hookrightarrow$ Unidirectional Finetuning & \xmark & 28.7 & 31.4 & $\approx \log_2 n \ll 10$ \\
\quad\quad\quad $\hookrightarrow$ Bidirectional Finetuning & \cmark & 28.1 & 28.6 & $\approx \log_2 n \ll 10$ \\
\bottomrule
\end{tabular}%
}
\caption{WMT English $\leftrightarrow$ German newstest2014 BLEU. Models capable of translating in both directions are marked with $\leftrightarrow$.}
\label{tab:results-newstest2014}
\end{table}

We first apply KERMIT on the competitive WMT 2014 English $\leftrightarrow$ German translation task.
We follow the hyperparameter settings of the base Transformer \citep{tensor2tensor}. However, since KERMIT does not have an encoder, we simply double the decoder width. We perform no additional hyperparameter tuning. We also follow prior work \citep{gu-iclr-2018,stern-neurips-2018,stern-icml-2019,lee-emnlp-2018} in using distillation \citep{hinton-nips-2015,kim-emnlp-2016} to train our models. We follow \cite{stern-icml-2019} in using a balanced binary tree loss, and we similarly observe an empirically logarithmic number of generation steps in sequence length when using parallel decoding. However, unlike \cite{stern-icml-2019} we did not need to tune an EOS penalty, but simply set it to zero for all experiments.

We train several different KERMIT models for translation. First we train two unidirectional models, where the model observes a full source sentence (i.e., English or German) and is asked to generate the corresponding target sentence (i.e., German or English). These separately learn the conditional distributions $p(y \mid x)$ and $p(x \mid y)$, mimicking the traditional conditional generation setup.
On the WMT 2014 test set, we achieve 27.8/30.7 BLEU with this approach, roughly matching our base Transformer baseline of 27.8/31.2 BLEU.
We also train a bidirectional model on the union of the two unidirectional training sets, yielding a single model that captures both conditional distributions $p(y \mid x)$ and $p(x \mid y)$. We do not change any hyperparameters when training this model (i.e., we do not increase model capacity).
The combined approach obtains 27.2/27.6 BLEU, nearly matching the baseline for English $\to$ German but falling slightly behind in the reverse direction.

We also train a full joint model that captures the full joint distribution $p(x, y)$ and factorizations thereof. Like the bidirectional model, the joint model can translate in either direction, but it can additionally be used for sampling or completing partial inputs. We use the same hyperparameter set as before. Since the model is now faced with a much more challenging task, it does slightly worse when limited to the same model size, but still reaches a respectable 25.6/27.4 BLEU. Unlike the previous models, however, we can incorporate monolingual data into the joint model's training setup to supplement its knowledge of the marginals $p(x)$ and $p(y)$. We accordingly train a joint model with all our paired data and 1M additional samples of English and German monolingual data randomly selected from the WMT 2014 monolingual corpus. Without altering model capacity, we find that refining the marginals gives us a 1.2 BLEU improvement on German $\rightarrow$ English.
Finally, we take the model which was trained on the full joint distribution with marginal refinement, and further finetune it on both the unidirectional and bidirectional settings. We find a small improvement in BLEU over the original models in both settings.

Table \ref{tab:results-newstest2014} summarizes our results. We emphasize that virtually all of our models outperform prior non-fully-autoregressive approaches in terms of BLEU. We also note that the observed number of iterations required to generate $n$ tokens is roughly $\log_2 n$ due to the use of a balanced binary tree loss and parallel decoding, which is substantially lower than autoregressive models which require $n$ steps. Some examples of parallel decodes are shown in Figure~\ref{fig:mt_btree_decodes}. Our models require an average of 5.5-6.5 decoding iterations for the sentences in the test set, outperforming the constant-time models of \cite{lee-emnlp-2018} which require 10 iterations in both BLEU and empirical decoding complexity.

We also draw samples from the model to highlight its infilling and generation capabilities. Figure \ref{fig:mt_samples} captures some examples. We first show unconditional sampling of an (English, German) sentence pair. We also take a translation example from the newstest2013 dev set and split it in half, sampling completions after seeding the English side with the first half and the German side with the second half. We find the model is capable of generating a very diverse set of coherent samples.


\begin{figure}[t]
\small
\begin{flushleft}
\textbf{Input:} In order to develop such a concept, the town is reliant on the cooperation of its citizens. \\
\vspace{0.5em}
\textbf{Predicted:} Um ein solches Konzept zu entwickeln, ist die Stadt auf die Zusammenarbeit ihrer Bürger angewiesen. \\
\vspace{0.5em}
\textbf{Parallel decode:}
\end{flushleft}
\vspace{-1.5em}
\begin{center}
\resizebox{\textwidth}{!}{%
\begin{tikzpicture}[every node/.style={anchor=base,inner sep=0.1em}]
\node[gray!60] (node-0-0) {Um\_};
\node[gray!60,base right=0cm of node-0-0] (node-0-1) {ein\_};
\node[gray!60,base right=0cm of node-0-1] (node-0-2) {solches\_};
\node[gray!60,base right=0cm of node-0-2] (node-0-3) {Konzept\_};
\node[gray!60,base right=0cm of node-0-3] (node-0-4) {zu\_};
\node[gray!60,base right=0cm of node-0-4] (node-0-5) {entwickeln\_};
\node[gray!60,base right=0cm of node-0-5] (node-0-6) {, \_};
\node[gray!60,base right=0cm of node-0-6] (node-0-7) {ist\_};
\node[blue,base right=0cm of node-0-7] (node-0-8) {\uline{die\_}};
\node[gray!60,base right=0cm of node-0-8] (node-0-9) {Stadt\_};
\node[gray!60,base right=0cm of node-0-9] (node-0-10) {auf\_};
\node[gray!60,base right=0cm of node-0-10] (node-0-11) {die\_};
\node[gray!60,base right=0cm of node-0-11] (node-0-12) {Zusammenarbeit\_};
\node[gray!60,base right=0cm of node-0-12] (node-0-13) {ihrer\_};
\node[gray!60,base right=0cm of node-0-13] (node-0-14) {Bürger\_};
\node[gray!60,base right=0cm of node-0-14] (node-0-15) {angewiesen\_};
\node[gray!60,base right=0cm of node-0-15] (node-0-16) {.\_};
\node[gray!60,below=0.1cm of node-0-0] (node-1-0) {Um\_};
\node[gray!60,base right=0cm of node-1-0] (node-1-1) {ein\_};
\node[gray!60,base right=0cm of node-1-1] (node-1-2) {solches\_};
\node[blue,base right=0cm of node-1-2] (node-1-3) {\uline{Konzept\_}};
\node[gray!60,base right=0cm of node-1-3] (node-1-4) {zu\_};
\node[gray!60,base right=0cm of node-1-4] (node-1-5) {entwickeln\_};
\node[gray!60,base right=0cm of node-1-5] (node-1-6) {, \_};
\node[gray!60,base right=0cm of node-1-6] (node-1-7) {ist\_};
\node[black,base right=0cm of node-1-7] (node-1-8) {die\_};
\node[gray!60,base right=0cm of node-1-8] (node-1-9) {Stadt\_};
\node[gray!60,base right=0cm of node-1-9] (node-1-10) {auf\_};
\node[gray!60,base right=0cm of node-1-10] (node-1-11) {die\_};
\node[blue,base right=0cm of node-1-11] (node-1-12) {\uline{Zusammenarbeit\_}};
\node[gray!60,base right=0cm of node-1-12] (node-1-13) {ihrer\_};
\node[gray!60,base right=0cm of node-1-13] (node-1-14) {Bürger\_};
\node[gray!60,base right=0cm of node-1-14] (node-1-15) {angewiesen\_};
\node[gray!60,base right=0cm of node-1-15] (node-1-16) {.\_};
\node[gray!60,below=0.1cm of node-1-0] (node-2-0) {Um\_};
\node[blue,base right=0cm of node-2-0] (node-2-1) {\uline{ein\_}};
\node[gray!60,base right=0cm of node-2-1] (node-2-2) {solches\_};
\node[black,base right=0cm of node-2-2] (node-2-3) {Konzept\_};
\node[gray!60,base right=0cm of node-2-3] (node-2-4) {zu\_};
\node[gray!60,base right=0cm of node-2-4] (node-2-5) {entwickeln\_};
\node[blue,base right=0cm of node-2-5] (node-2-6) {\uline{, \_}};
\node[gray!60,base right=0cm of node-2-6] (node-2-7) {ist\_};
\node[black,base right=0cm of node-2-7] (node-2-8) {die\_};
\node[gray!60,base right=0cm of node-2-8] (node-2-9) {Stadt\_};
\node[blue,base right=0cm of node-2-9] (node-2-10) {\uline{auf\_}};
\node[gray!60,base right=0cm of node-2-10] (node-2-11) {die\_};
\node[black,base right=0cm of node-2-11] (node-2-12) {Zusammenarbeit\_};
\node[gray!60,base right=0cm of node-2-12] (node-2-13) {ihrer\_};
\node[blue,base right=0cm of node-2-13] (node-2-14) {\uline{Bürger\_}};
\node[gray!60,base right=0cm of node-2-14] (node-2-15) {angewiesen\_};
\node[gray!60,base right=0cm of node-2-15] (node-2-16) {.\_};
\node[blue,below=0.1cm of node-2-0] (node-3-0) {\uline{Um\_}};
\node[black,base right=0cm of node-3-0] (node-3-1) {ein\_};
\node[blue,base right=0cm of node-3-1] (node-3-2) {\uline{solches\_}};
\node[black,base right=0cm of node-3-2] (node-3-3) {Konzept\_};
\node[gray!60,base right=0cm of node-3-3] (node-3-4) {zu\_};
\node[blue,base right=0cm of node-3-4] (node-3-5) {\uline{entwickeln\_}};
\node[black,base right=0cm of node-3-5] (node-3-6) {, \_};
\node[blue,base right=0cm of node-3-6] (node-3-7) {\uline{ist\_}};
\node[black,base right=0cm of node-3-7] (node-3-8) {die\_};
\node[blue,base right=0cm of node-3-8] (node-3-9) {\uline{Stadt\_}};
\node[black,base right=0cm of node-3-9] (node-3-10) {auf\_};
\node[blue,base right=0cm of node-3-10] (node-3-11) {\uline{die\_}};
\node[black,base right=0cm of node-3-11] (node-3-12) {Zusammenarbeit\_};
\node[blue,base right=0cm of node-3-12] (node-3-13) {\uline{ihrer\_}};
\node[black,base right=0cm of node-3-13] (node-3-14) {Bürger\_};
\node[blue,base right=0cm of node-3-14] (node-3-15) {\uline{angewiesen\_}};
\node[gray!60,base right=0cm of node-3-15] (node-3-16) {.\_};
\node[black,below=0.1cm of node-3-0] (node-4-0) {Um\_};
\node[black,base right=0cm of node-4-0] (node-4-1) {ein\_};
\node[black,base right=0cm of node-4-1] (node-4-2) {solches\_};
\node[black,base right=0cm of node-4-2] (node-4-3) {Konzept\_};
\node[blue,base right=0cm of node-4-3] (node-4-4) {\uline{zu\_}};
\node[black,base right=0cm of node-4-4] (node-4-5) {entwickeln\_};
\node[black,base right=0cm of node-4-5] (node-4-6) {, \_};
\node[black,base right=0cm of node-4-6] (node-4-7) {ist\_};
\node[black,base right=0cm of node-4-7] (node-4-8) {die\_};
\node[black,base right=0cm of node-4-8] (node-4-9) {Stadt\_};
\node[black,base right=0cm of node-4-9] (node-4-10) {auf\_};
\node[black,base right=0cm of node-4-10] (node-4-11) {die\_};
\node[black,base right=0cm of node-4-11] (node-4-12) {Zusammenarbeit\_};
\node[black,base right=0cm of node-4-12] (node-4-13) {ihrer\_};
\node[black,base right=0cm of node-4-13] (node-4-14) {Bürger\_};
\node[black,base right=0cm of node-4-14] (node-4-15) {angewiesen\_};
\node[blue,base right=0cm of node-4-15] (node-4-16) {\uline{.\_}};
\end{tikzpicture}
}
\end{center}
\vspace{-0.5em}
\parbox{\textwidth}{\dotfill}
\begin{flushleft}
\textbf{Input:} Frühere Gespräche zwischen den Parteien haben nur wenig zur Beilegung der Spannungen beigetragen, die durch eine Reihe von Zusammenstößen in diesem Jahr befeuert wurden. \\
\vspace{0.5em}
\textbf{Predicted:} Previous talks between the parties have done little to resolve the tensions fueled by a series of clashes this year. \\
\vspace{0.5em}
\textbf{Parallel decode:}
\end{flushleft}
\vspace{-1.5em}
\begin{center}
\resizebox{\textwidth}{!}{%
\begin{tikzpicture}[every node/.style={anchor=base,inner sep=0.1em}]
\node[gray!60] (node-0-0) {Prev};
\node[gray!60,base right=0cm of node-0-0] (node-0-1) {ious\_};
\node[gray!60,base right=0cm of node-0-1] (node-0-2) {talks\_};
\node[gray!60,base right=0cm of node-0-2] (node-0-3) {between\_};
\node[gray!60,base right=0cm of node-0-3] (node-0-4) {the\_};
\node[gray!60,base right=0cm of node-0-4] (node-0-5) {parties\_};
\node[gray!60,base right=0cm of node-0-5] (node-0-6) {have\_};
\node[gray!60,base right=0cm of node-0-6] (node-0-7) {done\_};
\node[gray!60,base right=0cm of node-0-7] (node-0-8) {little\_};
\node[gray!60,base right=0cm of node-0-8] (node-0-9) {to\_};
\node[gray!60,base right=0cm of node-0-9] (node-0-10) {resolve\_};
\node[gray!60,base right=0cm of node-0-10] (node-0-11) {the\_};
\node[blue,base right=0cm of node-0-11] (node-0-12) {\uline{tensions\_}};
\node[gray!60,base right=0cm of node-0-12] (node-0-13) {fueled\_};
\node[gray!60,base right=0cm of node-0-13] (node-0-14) {by\_};
\node[gray!60,base right=0cm of node-0-14] (node-0-15) {a\_};
\node[gray!60,base right=0cm of node-0-15] (node-0-16) {series\_};
\node[gray!60,base right=0cm of node-0-16] (node-0-17) {of\_};
\node[gray!60,base right=0cm of node-0-17] (node-0-18) {cla};
\node[gray!60,base right=0cm of node-0-18] (node-0-19) {she};
\node[gray!60,base right=0cm of node-0-19] (node-0-20) {s\_};
\node[gray!60,base right=0cm of node-0-20] (node-0-21) {this\_};
\node[gray!60,base right=0cm of node-0-21] (node-0-22) {year\_};
\node[gray!60,base right=0cm of node-0-22] (node-0-23) {.\_};
\node[gray!60,below=0.1cm of node-0-0] (node-1-0) {Prev};
\node[gray!60,base right=0cm of node-1-0] (node-1-1) {ious\_};
\node[gray!60,base right=0cm of node-1-1] (node-1-2) {talks\_};
\node[gray!60,base right=0cm of node-1-2] (node-1-3) {between\_};
\node[gray!60,base right=0cm of node-1-3] (node-1-4) {the\_};
\node[gray!60,base right=0cm of node-1-4] (node-1-5) {parties\_};
\node[blue,base right=0cm of node-1-5] (node-1-6) {\uline{have\_}};
\node[gray!60,base right=0cm of node-1-6] (node-1-7) {done\_};
\node[gray!60,base right=0cm of node-1-7] (node-1-8) {little\_};
\node[gray!60,base right=0cm of node-1-8] (node-1-9) {to\_};
\node[gray!60,base right=0cm of node-1-9] (node-1-10) {resolve\_};
\node[gray!60,base right=0cm of node-1-10] (node-1-11) {the\_};
\node[black,base right=0cm of node-1-11] (node-1-12) {tensions\_};
\node[gray!60,base right=0cm of node-1-12] (node-1-13) {fueled\_};
\node[gray!60,base right=0cm of node-1-13] (node-1-14) {by\_};
\node[gray!60,base right=0cm of node-1-14] (node-1-15) {a\_};
\node[gray!60,base right=0cm of node-1-15] (node-1-16) {series\_};
\node[gray!60,base right=0cm of node-1-16] (node-1-17) {of\_};
\node[blue,base right=0cm of node-1-17] (node-1-18) {\uline{cla}};
\node[gray!60,base right=0cm of node-1-18] (node-1-19) {she};
\node[gray!60,base right=0cm of node-1-19] (node-1-20) {s\_};
\node[gray!60,base right=0cm of node-1-20] (node-1-21) {this\_};
\node[gray!60,base right=0cm of node-1-21] (node-1-22) {year\_};
\node[gray!60,base right=0cm of node-1-22] (node-1-23) {.\_};
\node[gray!60,below=0.1cm of node-1-0] (node-2-0) {Prev};
\node[gray!60,base right=0cm of node-2-0] (node-2-1) {ious\_};
\node[blue,base right=0cm of node-2-1] (node-2-2) {\uline{talks\_}};
\node[gray!60,base right=0cm of node-2-2] (node-2-3) {between\_};
\node[gray!60,base right=0cm of node-2-3] (node-2-4) {the\_};
\node[gray!60,base right=0cm of node-2-4] (node-2-5) {parties\_};
\node[black,base right=0cm of node-2-5] (node-2-6) {have\_};
\node[gray!60,base right=0cm of node-2-6] (node-2-7) {done\_};
\node[gray!60,base right=0cm of node-2-7] (node-2-8) {little\_};
\node[blue,base right=0cm of node-2-8] (node-2-9) {\uline{to\_}};
\node[gray!60,base right=0cm of node-2-9] (node-2-10) {resolve\_};
\node[gray!60,base right=0cm of node-2-10] (node-2-11) {the\_};
\node[black,base right=0cm of node-2-11] (node-2-12) {tensions\_};
\node[gray!60,base right=0cm of node-2-12] (node-2-13) {fueled\_};
\node[gray!60,base right=0cm of node-2-13] (node-2-14) {by\_};
\node[blue,base right=0cm of node-2-14] (node-2-15) {\uline{a\_}};
\node[gray!60,base right=0cm of node-2-15] (node-2-16) {series\_};
\node[gray!60,base right=0cm of node-2-16] (node-2-17) {of\_};
\node[black,base right=0cm of node-2-17] (node-2-18) {cla};
\node[gray!60,base right=0cm of node-2-18] (node-2-19) {she};
\node[gray!60,base right=0cm of node-2-19] (node-2-20) {s\_};
\node[blue,base right=0cm of node-2-20] (node-2-21) {\uline{this\_}};
\node[gray!60,base right=0cm of node-2-21] (node-2-22) {year\_};
\node[gray!60,base right=0cm of node-2-22] (node-2-23) {.\_};
\node[gray!60,below=0.1cm of node-2-0] (node-3-0) {Prev};
\node[blue,base right=0cm of node-3-0] (node-3-1) {\uline{ious\_}};
\node[black,base right=0cm of node-3-1] (node-3-2) {talks\_};
\node[gray!60,base right=0cm of node-3-2] (node-3-3) {between\_};
\node[blue,base right=0cm of node-3-3] (node-3-4) {\uline{the\_}};
\node[gray!60,base right=0cm of node-3-4] (node-3-5) {parties\_};
\node[black,base right=0cm of node-3-5] (node-3-6) {have\_};
\node[gray!60,base right=0cm of node-3-6] (node-3-7) {done\_};
\node[blue,base right=0cm of node-3-7] (node-3-8) {\uline{little\_}};
\node[black,base right=0cm of node-3-8] (node-3-9) {to\_};
\node[gray!60,base right=0cm of node-3-9] (node-3-10) {resolve\_};
\node[blue,base right=0cm of node-3-10] (node-3-11) {\uline{the\_}};
\node[black,base right=0cm of node-3-11] (node-3-12) {tensions\_};
\node[blue,base right=0cm of node-3-12] (node-3-13) {\uline{fueled\_}};
\node[gray!60,base right=0cm of node-3-13] (node-3-14) {by\_};
\node[black,base right=0cm of node-3-14] (node-3-15) {a\_};
\node[blue,base right=0cm of node-3-15] (node-3-16) {\uline{series\_}};
\node[gray!60,base right=0cm of node-3-16] (node-3-17) {of\_};
\node[black,base right=0cm of node-3-17] (node-3-18) {cla};
\node[gray!60,base right=0cm of node-3-18] (node-3-19) {she};
\node[blue,base right=0cm of node-3-19] (node-3-20) {\uline{s\_}};
\node[black,base right=0cm of node-3-20] (node-3-21) {this\_};
\node[gray!60,base right=0cm of node-3-21] (node-3-22) {year\_};
\node[blue,base right=0cm of node-3-22] (node-3-23) {\uline{.\_}};
\node[blue,below=0.1cm of node-3-0] (node-4-0) {\uline{Prev}};
\node[black,base right=0cm of node-4-0] (node-4-1) {ious\_};
\node[black,base right=0cm of node-4-1] (node-4-2) {talks\_};
\node[blue,base right=0cm of node-4-2] (node-4-3) {\uline{between\_}};
\node[black,base right=0cm of node-4-3] (node-4-4) {the\_};
\node[blue,base right=0cm of node-4-4] (node-4-5) {\uline{parties\_}};
\node[black,base right=0cm of node-4-5] (node-4-6) {have\_};
\node[blue,base right=0cm of node-4-6] (node-4-7) {\uline{done\_}};
\node[black,base right=0cm of node-4-7] (node-4-8) {little\_};
\node[black,base right=0cm of node-4-8] (node-4-9) {to\_};
\node[blue,base right=0cm of node-4-9] (node-4-10) {\uline{resolve\_}};
\node[black,base right=0cm of node-4-10] (node-4-11) {the\_};
\node[black,base right=0cm of node-4-11] (node-4-12) {tensions\_};
\node[black,base right=0cm of node-4-12] (node-4-13) {fueled\_};
\node[blue,base right=0cm of node-4-13] (node-4-14) {\uline{by\_}};
\node[black,base right=0cm of node-4-14] (node-4-15) {a\_};
\node[black,base right=0cm of node-4-15] (node-4-16) {series\_};
\node[blue,base right=0cm of node-4-16] (node-4-17) {\uline{of\_}};
\node[black,base right=0cm of node-4-17] (node-4-18) {cla};
\node[blue,base right=0cm of node-4-18] (node-4-19) {\uline{she}};
\node[black,base right=0cm of node-4-19] (node-4-20) {s\_};
\node[black,base right=0cm of node-4-20] (node-4-21) {this\_};
\node[blue,base right=0cm of node-4-21] (node-4-22) {\uline{year\_}};
\node[black,base right=0cm of node-4-22] (node-4-23) {.\_};
\end{tikzpicture}
}
\end{center}
\vspace{-0.5em}
\parbox{\textwidth}{\dotfill}
\caption{Example parallel decodes using KERMIT for English $\to$ German and German $\to$ English translation. In each row, the blue underlined tokens are those being inserted, and the gray tokens are those from the final output that have not yet been generated. Empirically, KERMIT requires only $\approx \log_2 n$ steps to generate a sequence of length $n$ when trained with a balanced binary tree prior.}
\label{fig:mt_btree_decodes}
\end{figure}

\begin{figure*}[t]
\small
\begin{flushleft}
\textbf{No seeding (unconditional):} \\
\vspace{0.5em}
\textbf{English:} Nonetheless, we feel, with fury, at the fact that the 500 million have no contradiction on a common approach to productivity. \\
\textbf{German:} Dennoch sind wir mit Wut der Ansicht, dass die 500 Millionen keinen Widerspruch in einem gemeinsamen Produktivitätsansatz aufweisen.
\end{flushleft}
\vspace{-0.25em}
\parbox{\textwidth}{\dotfill}
\begin{flushleft}
\textbf{English Groundtruth:} - Please tell us, in simple terms, about the work your research group does. \\
\textbf{German Groundtruth:} - Beschreiben Sie bitte kurz, welche Forschungen Ihre Gruppe betreibt. \\
\vspace{0.5em}
\textbf{English Seed:} \gray{- Please tell us, in simple terms} \\
\textbf{German Seed:} \gray{welche Forschungen Ihre Gruppe betreibt.} \\
\vspace{0.5em}
\textbf{English:} \gray{- Please tell us} \black{what sort of research your group is conducting}\gray{, in simple terms}\black{.} \\
\textbf{German:} \black{- Bitte teilen Sie uns einfach mit,} \gray{welche Forschungen Ihre Gruppe betreibt.} \\
\vspace{0.5em}
\textbf{English:} \gray{- Please tell us, in} \black{quite} \gray{simple terms}\black{, what kind of research your group operates.} \\
\textbf{German:} \black{- Bitte teilen Sie uns in ganz einfach mit,} \gray{welche Forschungen Ihre Gruppe betreibt.} \\
\vspace{0.5em}
\textbf{English:} \gray{- Please tell us} \black{what research}\gray{, in simple terms}\black{, what your group actually runs.} \\
\textbf{German: } \black{- Bitte sagen Sie uns ganz einfach,} \gray{welche Forschungen Ihre Gruppe} \black{eigentlich} \gray{betreibt.} \\
\vspace{0.5em}
\textbf{English: } \gray{- Please}\black{,} \gray{tell us} \black{what research your group is doing}\gray{, in} \black{more} \gray{simple terms}\black{.} \\
\textbf{German: } \black{- Bitte sagen Sie uns,} \gray{welche Forschungen Ihre Gruppe} \black{einfacher} \gray{betreibt.} \\
\vspace{0.5em}
\textbf{English: } \gray{- Please tell us} \black{what your group will be conducting public research on}\gray{, in simple terms}\black{.} \\
\textbf{German: } \black{- Bitte teilen Sie uns einfach mit,} \gray{welche Forschungen Ihre Gruppe betreibt.} \\
\vspace{0.5em}
\textbf{English: } \gray{- Please tell us,} \black{what sort of research your group is undertaking} \gray{in simple terms}\black{.} \\
\textbf{German: } \black{- Bitte sagen Sie uns,} \gray{welche Forschungen Ihre Gruppe betreibt.}\end{flushleft}
\vspace{-0.25em}
\parbox{\textwidth}{\dotfill}
\caption{Paired translation samples drawn from KERMIT, with and without seeding the initial canvas with text. In the bottom portion of the figure, the seed text is shown in gray, and different continuations sampled from the model are shown in black. We emphasize the diversity of generation.}
\label{fig:mt_samples}
\end{figure*}

%

\subsection{Representation Learning}
\label{sec:representationlearning}

\begin{table}[t]
\centering
\resizebox{\textwidth}{!}{
\begin{tabular}{lcccccccccccc}
\toprule
\bfseries Model &
\bfseries Generative?
& \bfseries  CoLA & \bfseries SST-2 & \bfseries MRPC & \bfseries STS-B & \bfseries QQP & \bfseries MNLI-(m/mm) & \bfseries QNLI & \bfseries RTE & \bfseries WNLI & \bfseries AX & \bfseries Score  \\
\midrule
GPT \citep{radford-openai-2018} &\cmark & 45.4 & 91.3 & 82.3/75.7 & 82.0/80.0 & 70.3/88.5 & 82.1/81.4 & 87.4 & 56.0 & 53.4 & 29.8 & 72.8 \\
BERT \citep{devlin-naacl-2019} & \xmark & 60.5 & 94.9 & 89.3/85.4 & 87.6/86.5 & 72.1/89.3 & 86.7/85.9 & 92.7 & 70.1 & 65.1 & 39.6 & 80.5 \\
\midrule
KERMIT & \cmark & 60.0 & 94.2 & 88.6/84.3 & 86.6/85.6 & 71.7/89.0 & 85.6/85.2 & 92.0 & 68.4 & 65.1 & 37.6 & 79.8 \\
\bottomrule
\end{tabular}%
}
\caption{GLUE benchmark scores (as computed by the GLUE evaluation server). Of these models, only GPT and KERMIT admit a straightforward generation process.}
\label{tab:glue}
\end{table}

Like its close friend BERT \citep{devlin-naacl-2019}, KERMIT can also be used for self-supervised representation learning and applied to various language understanding tasks. We follow the same training procedure and hyperparameter setup as BERT$_\textsc{large}$. However, instead of masking 15\% of the tokens and replacing them with blank tokens like in BERT \citep{devlin-naacl-2019}, KERMIT simply drops them out completely from the sequence.

Prior to BERT, the best representation learning approach was to use a language model such as GPT \citep{radford-openai-2018}. BERT outperforms GPT in large part because of its \emph{deeply bi-directional} architecture, but in the process BERT sacrifices the ability to perform straightforward generation. While we find KERMIT to perform slightly behind BERT, KERMIT maintains the ability to generate text while obtaining results that are much closer to BERT rather than GPT. The GLUE benchmark \citep{wang-iclr-2019} results are summarized in Table \ref{tab:glue}.

\subsection{Zero-Shot Cloze Question Answering}
\label{sec:clozeqa}


Finally, we also investigate the infilling abilities of KERMIT and related approaches by evaluating their performance on zero-shot cloze question answering. In particular, we aim to understand how effective these models are for fill-in-the-blank-style question answering after being trained only on language modeling data without any task-specific fine-tuning.

For this experiment, we use the human-annotated QA2D dataset assembled by \citet{demszky-arxiv-2018}, which consists of examples from the SQuAD dataset \citep{rajpurkar-emnlp-2016} in which the answer has been extended from a single phrase into a full declarative sentence. These can be transformed into cloze instances by removing the answer phrase from the declarative output. For example, given the question ``When was Madonna born?'' and the answer ``August 16, 1958'', the full declarative answer would be ``Madonna was born on August 16, 1958.'', and the associated cloze instance would be ``Madonna was born on $\rule{1cm}{0.15mm}$ .''

We take the KERMIT model trained from Section \ref{sec:representationlearning} and two powerful language models (BERT \citep{devlin-naacl-2019} and the largest public version of GPT-2\footnote{The 345M parameter ``medium size'' model.} \citep{radford-openai-2019}), and evaluate their ability to fill in the blank of each cloze instance, each without specifically being trained on data of this form. We employ different decoding strategies as required for each model, detailed below.

\paragraph{KERMIT} We split the passage in half and present KERMIT with examples of the form $$\texttt{\small [CLS] passage(1/2) [SEP] passage(2/2) question cloze(left) cloze(right) [SEP]}$$ where \texttt{\small cloze(left)} and \texttt{\small cloze(right)} are the portions of the declarative answer before and after the gap. Since KERMIT can natively perform insertions, we simply perform a parallel decode constrained to take place within the gap and extract the output as our answer.

\paragraph{BERT} We split the passage in half and present BERT with examples of the form $$\texttt{\small [CLS] passage(1/2) [SEP] passage(2/2) question cloze(left) [MASK]*n cloze(right) [SEP]}$$ Here we include explicit \texttt{\scriptsize[MASK]} tokens, running separate decodes with $n = 1, 2, \dots$ up to $4$ or the oracle answer length, whichever is greater. We then choose the one with the highest score under the model and extract the outputs at the masked positions as the answer. Each decode consists of a beam search in which one \texttt{\small [MASK]} is filled at a time. For each element on the beam, we choose the remaining \texttt{\small [MASK]} position with the highest confidence (lowest entropy) as the next position to fill. We found that this beam-search did substantially better than left-to-right decoding or parallel decoding.

\paragraph{GPT-2} For GPT-2, a left-to-right language model, we cannot directly condition on both the left and right context. Instead, we first present the model with the prefix $$\texttt{\small passage question cloze(left)}$$ and sample continuations of varying lengths. For each continuation, we then append \texttt{\small cloze(right)} and compute the score of the full sequence under the model. We select the best-scoring sequence and extract the portion in the gap as the answer. To efficiently obtain continuations of varying lengths, we generate 20 extended continuations from the model, then treat all prefixes of those continuations as candidate values to go in the gap.

\begin{figure}[t]
\begin{minipage}{.5\textwidth}
\scriptsize
\begin{flushleft}
Plymouth has a post-war shopping area in the city centre with substantial pedestrianisation. At the west end of the zone inside a grade II listed building is the Pannier Market that was completed in 1959 – pannier meaning "basket" from French, so it translates as "basket market". In terms of retail floorspace, Plymouth is ranked in the top five in the South West, and 29th nationally. Plymouth was one of the first ten British cities to trial the new Business Improvement District initiative. The Tinside Pool is situated at the foot of the Hoe and became a grade II listed building in 1998 before being restored to its 1930s look for £3.4 million. \textbf{What notable location was named a grade II listed building in 1998? \_\_\_ was named a grade II listed building in 1998 .}
\end{flushleft}
\end{minipage}
\begin{minipage}{.5\textwidth}
\begin{tabular}{ll}
\bfseries Model                     & \bfseries Answer  \\
\midrule
GPT-2                               & $\rightarrow$ ``Plymouth''          \\
{\color{gray} + Oracle Length} & $\rightarrow$ ``A listed building'' \\
\midrule
BERT                                & $\rightarrow$ ``plymouth''          \\
{\color{gray} + Oracle Length} & $\rightarrow$ ``: the pool''        \\
\midrule
KERMIT                              & $\rightarrow$ \textbf{``the tinside pool''}  \\
\midrule
Correct                             & $\rightarrow$ ``Tinside Pool''
\end{tabular}%
\end{minipage}%
\vspace{0.5em}
\parbox{\textwidth}{\dotfill}
\caption{Example of KERMIT, BERT and GPT-2 performing zero-shot cloze question answering on SQuAD. The question and cloze question are bolded. Note that BERT and GPT-2 prefer a shorter, incorrect answer, unless given the oracle answer length.}
\label{fig:clozeqa_example}
\end{figure}

\begin{wraptable}[10]{r}{0.4\linewidth}
\vspace{-1em}
\small
\centering
\resizebox{0.4\columnwidth}{!}{
\begin{tabular}{lcc}
\toprule
\bfseries Model & \bfseries Exact Match & \bfseries F1 \\
\midrule
GPT-2 & 10.9 & 16.6 \\
{\color{gray}\quad + Oracle Length} & \color{gray}12.2 & \color{gray}18.2 \\
BERT & 12.3 & 18.9 \\
{\color{gray}\quad + Oracle Length} & \color{gray}16.2 & \color{gray}23.1 \\
\midrule
KERMIT & \textbf{20.9} & \textbf{30.3} \\
\bottomrule
\end{tabular}%
}
\caption{SQuAD zero-shot cloze question answering.}
\label{tab:cloze-results}
\end{wraptable}

We evaluate on 50,000 cloze-formulated questions from SQuAD, using the standard SQuAD evaluation script to compute accuracy in terms of exact match and token-level F1. Results are presented in Table~\ref{tab:cloze-results}.
KERMIT performs significantly better on this zero-shot cloze task than the other two approaches thanks to its infilling capabilities learned through its insertion-oriented objective, achieving 30.3 F1 and 20.9\% exact match. BERT's performance falls short of KERMIT, as it often prefers shorter completions since it is not required to handle length modeling during training. GPT-2 lags further behind the others due to its inability to condition on the context on both sides of the gap during inference. Even when the oracle length (i.e., the ground-truth length of the answer) is provided to BERT and GPT-2, KERMIT still substantially outperforms all other models.

\section{Conclusion}
%
In this paper, we present KERMIT, an insertion-based framework for sequences that can model the joint data distribution and its decompositions (i.e., marginals and conditionals). KERMIT can generate text in an arbitrary order -- including bidirectional machine translation and cloze-style infilling -- and empirically can generate sequences in logarithmic time. It uses a simple neural architecture that can additionally produce contextualized vector representations of words and sentences. We find KERMIT is capable of matching or exceeding state-of-the-art performance on three diverse tasks: machine translation, representation learning, and zero shot cloze question answering.

\subsubsection*{Acknowledgments}
We give thanks to Samy Bengio, Zhifeng Chen, Jamie Kiros, Luheng He, Geoffrey Hinton, Quoc Le, Lala Li, Mohammad Norouzi, Yu Zhang, and the Google Brain team for useful discussions and technical assistance. Special thanks to Jamie Kiros for brainstorming the name KERMIT.


\bibliography{paper}
\bibliographystyle{apalike}

\end{document}